\documentclass{article}
\usepackage{spconf,amsmath,graphicx}

\usepackage[inline]{enumitem}
\usepackage{booktabs} 

\setlist{nosep, leftmargin=14pt}

\usepackage{mwe} 

\newcommand{\cvmail}{{CVMAIL\_$\times$\_MIHL}\xspace}
\newcommand{\coolpeace}{{Cool\_Peace}\xspace}
\newcommand{\viu}{{VIU}\xspace}
\newcommand{\zuang}{{zuang}\xspace}

\title{Overview of the CXR-LT 2026 Challenge: Multi-Center Long-Tailed and Zero Shot Chest X-ray Classification}

\name{\begin{tabular}{c}
Hexin Dong$^{1}$\sthanks{These authors contributed equally to this work.}, Yi Lin$^{1\star}$, Pengyu Zhou$^2$, Xuan Zhong Feng$^1$, Alan Clint Legasto$^1$\\ Mingquan Lin$^3$, Hao Chen$^4$, Yuzhe Yang$^5$, George Shih$^1$, Yifan Peng$^{1}$\sthanks{Corresponding author. Email: yip4002@med.cornell.edu}
\end{tabular}}
\address{$^1$Weill Cornell Medicine;
$^2$Peking Union Medical College;
$^3$University of Minnesota;\\
$^4$Hong Kong University of Science and Technology;\\
$^5$University of California, Los Angeles\\}

\begin{document}
\ninept

\maketitle
\begin{abstract}
Chest X-ray (CXR) interpretation is hindered by the long-tailed distribution of pathologies and the open-world nature of clinical environments. Existing benchmarks often rely on closed-set classes from single institutions, failing to capture the prevalence of rare diseases or the appearance of novel findings. To address this, we present the \textbf{CXR-LT 2026} challenge. This third iteration of the benchmark introduces a multi-center dataset comprising over 145,000 images from PadChest and NIH Chest X-ray datasets. The challenge defines two core tasks: (1) Robust Multi-Label Classification on 30 known classes and (2) Open-World Generalization to 6 unseen (out-of-distribution) rare disease classes. We report the results of the top-performing teams, evaluating them via mean Average Precision (mAP), AUROC, and F1-score. The winning solutions achieved an mAP of 0.5854 on Task 1 and 0.4315 on Task 2, demonstrating that large-scale vision-language pre-training significantly mitigates the performance drop typically associated with zero-shot diagnosis.
\end{abstract}
\section{Introduction}
\label{sec:intro}

Chest X-rays (CXRs), like many other diagnostic medical exams, produce a long-tailed distribution of clinical findings, where a small subset of diseases occur frequently. Still, the vast majority are relatively rare ~\cite{zhou2021review}. Such an imbalance poses a challenge for standard deep learning methods, which exhibit bias toward the most common classes at the expense of the important but rare ``tail" classes ~\cite{holste2022long}. Although many existing methods~\cite{zhang2023deep} have been proposed to tackle this specific type of imbalance, only recently has attention been given to long-tailed medical image recognition problems ~\cite{zhang2021mbnm,yang2022proco,ju2021relational}.

To advance research on long-tailed learning in CXR analysis, we established the CXR-LT challenge series. The first event, CXR-LT 2023~\cite{holste2023cxr} provided high-quality benchmark data for model development and conducting comprehensive evaluations to identify ongoing issues impacting lung disease classification performance. Building on its success, CXR-LT 2024 ~\cite{lin2025cxr} significantly expanded the dataset to 377,110 CXRs annotated with 45 diseases, including 19 newly introduced rare disease findings. It also introduces a new zero-shot learning track to address key limitations identified in the previous edition.

In CXR-LT 2026, we continue and extend the benchmark by introducing three key changes. First, we leverage a new data source, PadChest~\cite{bustos2020padchest}, which has not previously been used in the CXR-LT benchmark and is publicly available, addressing recent access restrictions on MIMIC-CXR~\cite{PhysioNet-mimic-cxr-2.0.0}. Second, we introduce newly curated test sets annotated by three radiologists to provide higher-quality and more reliable evaluation. Third, unlike CXR-LT 2024, where all labels were extracted from free-text radiology reports, the development and test sets in CXR-LT 2026 use image-level annotations from radiologists, while the larger training set still relies on report-derived labels. This paper summarizes an overview of the challenge design, dataset characteristics, and the results of the top-performing teams.

\section{Materials and Methods}
\label{sec:method}

\subsection{Dataset Collection}

The CXR-LT 2026 dataset consists of over \textbf{145,000 chest X-rays} collected from two sources (Table~\ref{tab:cxrlt2026_dataset}):
\begin{enumerate*}[label=(\arabic*)]
    \item \textbf{PadChest}~\cite{bustos2020padchest}, which serves as the primary source of common chest diseases and establishes the base distribution of seen classes; and
    \item \textbf{NIH Chest X-ray}~\cite{summers2019nih}, which is incorporated to introduce domain shift and additional rare findings not present in PadChest.
\end{enumerate*}
By combining these two datasets, CXR-LT 2026 captures both long-tailed class imbalance and cross-domain variability, better reflecting the complexity of real-world clinical deployment scenarios.

\subsection{Label Collection and Task Definition}

\begin{figure*}[htb]
    \centering
    \includegraphics[width=\linewidth]{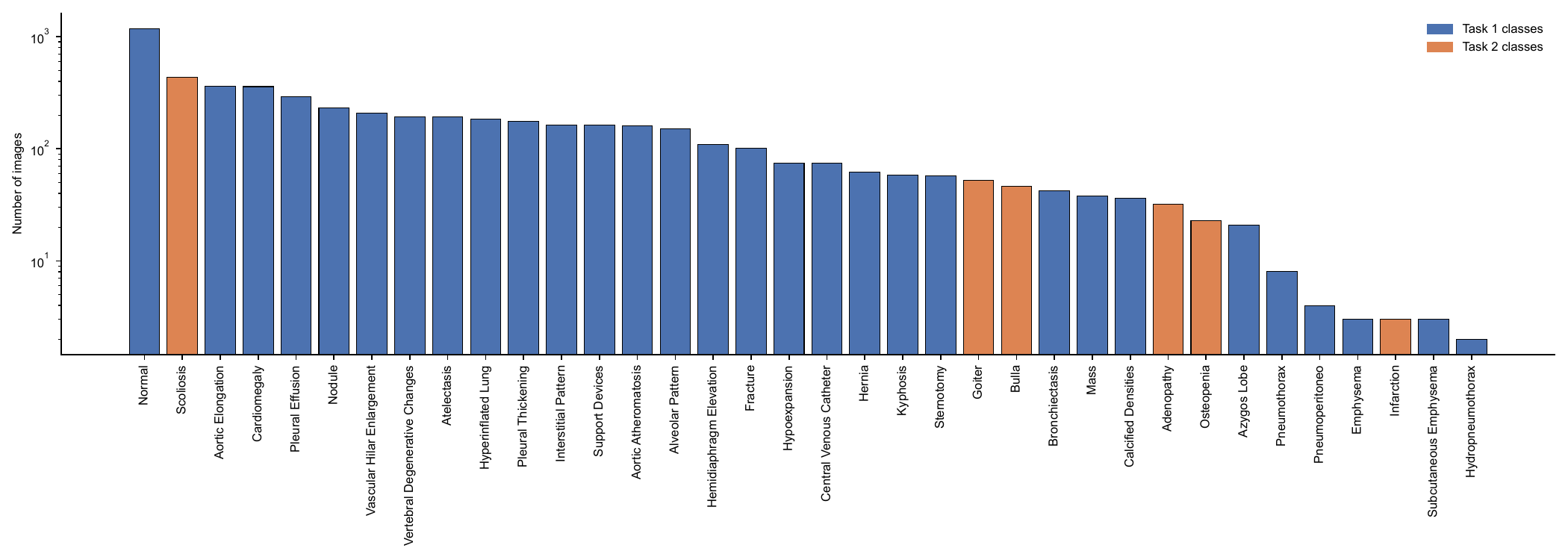}
    \caption{Per-class label distribution in the CXR-LT 2026 evaluation sets (development + test).}
    \label{fig:label_counts}
\end{figure*}

\subsubsection{Task 1: Robust Multi-Label Classification}

This task assesses the model's ability to classify \emph{30 ``Seen'' classes}. Following the setting in CXR-LT 2024 challenge ~\cite{lin2025cxr}, we first selected 20 categories that overlapped with the previous edition.
We then added 10 additional high-prevalence categories from PadChest-GR~\cite{de2025padchest} to better cover common clinical findings in CXR.
The final label set includes:
\begin{enumerate*}[label=(\arabic*)]
\item Alveolar Pattern,
\item Aortic Atheromatosis,
\item Aortic Elongation,
\item Atelectasis,
\item Azygos Lobe,
\item Bronchiectasis, 
\item Cardiomegaly,
\item Calcified Densities,
\item Central Venous Catheter, 
\item Emphysema,
\item Fracture,
\item Hemidiaphragm Elevation, 
\item Hernia,
\item Hydropneumothorax,
\item Hyperinflated Lung, 
\item Hypoexpansion, 
\item Interstitial Pattern, 
\item Kyphosis,
\item Mass,
\item Nodule,
\item Normal,
\item Pleural Effusion,
\item Pleural Thickening,
\item Pneumothorax,
\item Pneumoperitoneo,
\item Sternotomy,
\item Subcutaneous Emphysema,
\item Support Devices,
\item Vascular Hilar Enlargement, 
\item Vertebral Degenerative Changes.
\end{enumerate*}
Figure \ref{fig:label_counts} shows the label distribution in the evaluation set.
%

Specifically, the labels for both the development and test sets were derived from the PadChest-GR dataset~\cite{de2025padchest}, which is a radiologist-annotated subset of PadChest. In PadChest-GR, several radiologists provided image-level annotations for each chest X-ray. From this dataset, we selected 3,020 frontal chest X-rays (excluding lateral projections) that contained at least one of the 30 ``seen'' classes. These cases were then split into development and test sets at approximately a 1:1 ratio. Stratified sampling was applied to maintain the class distribution as evenly as possible across two splits, thereby ensuring a balanced and reliable evaluation setting.

The training set for Task~1 was constructed from the 142,928 images from the PadChest that were not included in the PadChest-GR subset. These images are accompanied by Spanish radiology reports, from which labels were automatically extracted. We used these report-derived labels as supervision for model training. In contrast, the radiologist-annotated PadChest-GR subset was reserved exclusively for development and final evaluation. 

This design allows models to learn from large-scale but potentially noisy labels, while being evaluated on higher-quality expert annotations. Such a separation better reflects realistic clinical AI development, where large quantities of weakly labeled data are available for training, but rigorous evaluation relies on carefully curated expert labels.

\begin{table}[t]
\centering
\renewcommand{\tabcolsep}{3pt}
\footnotesize
\caption{Overview of the CXR-LT 2026 dataset.}
\label{tab:cxrlt2026_dataset}
\begin{tabular}{cllrl}
\toprule
\textbf{Tasks}& \textbf{Dataset}& \textbf{Source} & \textbf{\# CXRs} & \textbf{Curation}  \\
\midrule
1 &  Training & PadChest & 142,928 & From reports\\
  &  Development & PadChest-GR & 1,620 & By radiologists\\
  &  Test & PadChest-GR & 1,400 & By radiologists\\
\midrule
2 &  Training & PadChest & 142,928 & From reports\\
  &  Development & PadChest-GR & 200 & By radiologists\\
  &  Internal Test & PadChest-GR & 305 & By radiologists\\
  &  External Test & NIH Chest X-ray & 79 & By radiologists\\
\bottomrule
\end{tabular}
\end{table}

\subsubsection{Task 2: Open-World Generalization}

This task assesses the model's ability to recognize \textbf{6 ``Unseen'' classes} that were strictly excluded from training. 
Labels for Task~2 were also derived from PadChest-GR. Building on the out-of-distribution (OOD) taxonomy defined in Task~3 of the CXR-LT 2024 challenge~\cite{lin2025cxr}, we retained the same five OOD categories from the previous edition to ensure continuity and comparability across challenges.
In addition, we introduced one additional rare condition, Goiter, resulting in a six-class OOD label set for open-world evaluation:
\begin{enumerate*}[label=(\arabic*)]
\item Adenopathy,
\item Bulla, 
\item Goiter,
\item Infarction, 
\item Osteopenia, 
\item Scoliosis.
\end{enumerate*}
Figure~\ref{fig:label_counts} shows the label distribution in the evaluation set.

We used the same training set as Task~1 for in-distribution training , without providing any labels for the six unseen classes.
To build an internal test set, we collected 505 chest X-ray images from PadChest-GR, which contained at least one of the 6 `unseen' classes. These images were split into development and test sets using an approximate 2:3 ratio, following the same protocol as Task~1.

In addition to the internal test set, we constructed an external test dataset using the NIH ChestX-ray dataset to further assess generalization. We selected 159 candidate chest X-rays with potential presence of at least one of the six ``unseen'' classes. All candidate images were then independently reviewed by three radiologists. Only those images for which at least two radiologists confirmed the corresponding finding were included in the final external test set, resulting in 79 externally validated OOD cases.

\subsection{Evaluation Metrics}

Since this is a multi-label classification problem with severe class imbalance, we used mean Average Precision (mAP) as the primary evaluation metric \cite{everingham2010pascal}, computed as the macro-averaged AP across the 30 classes in Task 1 or the 6 classes in Task~2. 

For completeness, we also reported mean AUC (mAUC) and mean F1 score (mF1), using a fixed decision threshold of 0.5 for each class. 
These metrics were displayed on the leaderboard to provide additional insights, but did not contribute to team rankings.

\subsection{Challenge procedure}

The challenge was conducted in a staged timeline to ensure fair evaluation and encourage iterative model development. The training set was released on November 24, 2025, allowing teams to explore the data, design models, and develop baseline approaches. After an initial period, the development set was released from December 1, 2025, to January 19, 2026. Participants could submit predictions on this set and access immediate feedback via the online CondaBench platform~\cite{codabench}, enabling them to analyze performance, identify weaknesses, and refine their models.

Once the development phase concluded, the final test set was released from January 20, 2026, to February 4, 2026, for evaluation. The evaluation period was based on  'blind test' rules. Specifically, teams were asked to submit no more than five valid predictions without any access to ground-truth labels and \textbf{could not} see any results for the submission. The submitted results were evaluated centrally, and the final leaderboard rankings were computed based on the predefined primary metric (mAP), with secondary metrics such as mAUC and mF1 provided for additional insights and used only to break ties when teams have the same mAP score.

After the final evaluation stage, each participating team was required to submit a technical report summarizing its methodology.

\section{Results}

\subsection{Team participation across stages}

A total of 72 teams initially registered and obtained access to the training data. During the development phase, 23 teams actively submitted predictions to the public leaderboard, totaling 2,475 submissions, reflecting sustained engagement and iterative model refinement.
At the final evaluation stage, 21 teams submitted complete test-set predictions for official ranking. Following the announcement of final results, all eligible teams were required to submit a technical report describing their methodology. Only teams that completed this final submission were considered for inclusion in the challenge summary.
Overall, participation gradually decreased from registration to final submission, a trend commonly observed in competitive machine learning challenges. Nevertheless, strong engagement during the development phase and competitive performance at the final stage indicate sustained interest and high-quality methodological contributions from the leading teams.

\subsection{Task 1: Robust Prediction Results}

\begin{table}[t]
\centering
\caption{Task~1 leaderboard (in-distribution multi-label classification).}
\label{tab:task1}
\begin{tabular}{c l c c c}
\toprule
Rank & Team Name & mAP & AUROC & F1 \\
\midrule
1 & \cvmail & 0.5854 & 0.9259 & 0.3518 \\
2 & \coolpeace & 0.4827 & 0.9186 & 0.3162 \\
3 & \viu                 & 0.4599 & 0.8827 & 0.4504 \\
4 & Bibimbap-Bueno      & 0.4297 & 0.8753 & 0.2482 \\
5 & Nikhil Rao Sulake   & 0.3950 & 0.8591 & 0.0945 \\
6 & UGVIA team          & 0.2362 & 0.7756 & 0.2353 \\
7 & 2025110451          & 0.0614 & 0.4925 & 0.0000 \\
8 & mshamani            & 0.0580 & 0.5020 & 0.0264 \\
9 & laghaei             & 0.0580 & 0.5012 & 0.0302 \\
10 & uccaeid            & 0.0576 & 0.5001 & 0.0337 \\
\bottomrule
\end{tabular}
\end{table}

Table~\ref{tab:task1} summarizes the results for the top ten teams in Task~1. Team \textbf{\cvmail} achieved a clear lead with an mAP of 0.5854, outperforming the second-place team \textbf{\coolpeace} (mAP 0.4827), by over 0.10. Its high AUROC of 0.9259 further indicates strong ranking capability across the full range of decision thresholds. 

Interestingly, the third-place team \textbf{\viu} obtained a lower mAP (0.4599) and AUROC (0.8827) than the top two teams, but achieved the highest F1-score (0.4504). This suggests that their model operates at a more favorable decision threshold for binary predictions, despite weaker overall ranking performance. 

Beyond the top three, performance drops sharply: the sixth-ranked team attains an mAP of 0.2362, and teams ranked seventh to tenth remain close to chance level in AUROC. These results highlight both the difficulty of long-tailed multi-label classification under distribution shift and the importance of well-calibrated decision thresholds with strong representation learning in this setting.

\begin{table}[t]
\centering
\caption{Task~2 leaderboard (zero-shot / OOD multi-label classification).}
\label{tab:task2}
\begin{tabular}{c l c c c}
\toprule
Rank & Team Name & mAP & AUROC & F1 \\
\midrule
1 & \cvmail & 0.4315 & 0.8073 & 0.1183 \\
2 & \coolpeace    & 0.3106 & 0.6705 & 0.2025 \\
3 & \zuang         & 0.2235 & 0.5903 & 0.1742 \\
4 & 2025110451    & 0.1799 & 0.5275 & 0.0000 \\
5 & laghaei       & 0.1686 & 0.5032 & 0.2303 \\
5 & uccaeid       & 0.1686 & 0.5032 & 0.2303 \\
5 & mshamani      & 0.1686 & 0.5032 & 0.2303 \\
\bottomrule
\end{tabular}
\end{table}

\subsection{Task 2: Open-World / OOD Multi-Label Classification}

Table~\ref{tab:task2} summarizes the top-performing submissions in Task 2. Overall, performance is substantially lower than in Task~1, reflecting the difficulty of detecting previously unseen findings under multi-center distribution shift.

Team \textbf{\cvmail} again ranked first, achieving an mAP of 0.4315 and an AUROC of 0.8073. Although these scores are lower than their in-distribution performance, they indicate that the underlying representation is still able to rank OOD cases reasonably well in a zero-shot setting. The second-place team \textbf{\coolpeace} obtained an mAP of 0.3106 and AUROC of 0.6705, but achieved a higher F1-score (0.2025) than \textbf{\cvmail}. This suggests that their model operates at a more favorable decision threshold for binary OOD detection despite weaker overall ranking ability. The third-place team \textbf{\zuang} reached an intermediate regime with mAP of 0.2235 and F1 of 0.1742.

Beyond the top three teams, performance rapidly approaches chance level, with AUROC values close to 0.5 and highly unstable F1-scores. These results highlight that, while strong vision--language or foundation-model-based approaches can provide non-trivial zero-shot capability, robust open-world detection of rare conditions remains a challenging and unresolved problem in CXR analysis.

\section{Discussion}

\subsection{The Dominance of Foundation Models}
The consistent top performance of \textbf{\cvmail} across both tasks suggests the use of a robust Vision-Language Pre-training (VLP) foundation model. Their ability to maintain high mAP on both seen and unseen classes supports the hypothesis that aligning visual features with medical text embeddings can substantially enhance generalization. In particular, strong performance in the open-world setting indicates that multimodal alignment enables meaningful semantic transfer, making it a highly effective strategy for developing clinically robust AI systems under distribution shift and limited supervision.

\subsection{Calibration vs. Discrimination}
A key observation in Task~1 is the mismatch between ranking-based metrics (mAP/AUROC) and threshold-dependent decision metrics (F1-score). The winning team \textbf{\cvmail} achieved the best mAP (0.5854) and AUROC (0.9259), indicating a strong ability to correctly rank positive cases across classes. In contrast, the third-place team \textbf{\viu} obtained lower mAP (0.4599) and AUROC (0.8827), but achieved the highest F1-score (0.4504). This suggests that their operating threshold was more favorable for binary decision-making, despite comparatively weaker overall ranking performance. 

These results highlight that strong discrimination capability does not automatically translate into optimal decision utilityat a fixed threshold. Therefore, both ranking metrics and threshold-dependent metrics, such as F1-score, should be considered when assessing models for potential clinical use.




\section{Conclusion}

CXR-LT 2026 provides a practical benchmark for long-tailed and zero-shot chest X-ray classification. Our results show that current AI models can already recognize a wide range of findings, but they still struggle to make reliable decisions on rare conditions and under limited supervision. We hope that this dataset and the released baseline solutions will support the community in building more robust and trustworthy clinical AI tools for chest X-ray interpretation.

\section{Acknowledgments}
\label{sec:acknowledgments}

This work was supported by the US National Science Foundation Career Grant (2145640). We thank Dr. Fengnian Zhao (West China School of Medicine) for his valuable assistance in supporting the evaluation phase of the challenge.


\begin{thebibliography}{10}

\bibitem{zhou2021review}
S~Kevin Zhou, Hayit Greenspan, Christos Davatzikos, James~S Duncan, Bram Van~Ginneken, Anant Madabhushi, Jerry~L Prince, Daniel Rueckert, and Ronald~M Summers,
\newblock ``A review of deep learning in medical imaging: Imaging traits, technology trends, case studies with progress highlights, and future promises,''
\newblock {\em Proceedings of the IEEE}, vol. 109, no. 5, pp. 820--838, 2021.

\bibitem{holste2022long}
Gregory Holste, Song Wang, Ziyu Jiang, Thomas~C Shen, George Shih, Ronald~M Summers, Yifan Peng, and Zhangyang Wang,
\newblock ``Long-tailed classification of thorax diseases on chest x-ray: A new benchmark study,''
\newblock in {\em MICCAI Workshop on Data Augmentation, Labelling, and Imperfections}. Springer, 2022, pp. 22--32.

\bibitem{zhang2023deep}
Yifan Zhang, Bingyi Kang, Bryan Hooi, Shuicheng Yan, and Jiashi Feng,
\newblock ``Deep long-tailed learning: A survey,''
\newblock {\em IEEE transactions on pattern analysis and machine intelligence}, vol. 45, no. 9, pp. 10795--10816, 2023.

\bibitem{zhang2021mbnm}
Ruru Zhang, E~Haihong, Lifei Yuan, Jiawen He, Hongxing Zhang, Shengjuan Zhang, Yanhui Wang, Meina Song, and Lifei Wang,
\newblock ``Mbnm: multi-branch network based on memory features for long-tailed medical image recognition,''
\newblock {\em Computer Methods and Programs in Biomedicine}, vol. 212, pp. 106448, 2021.

\bibitem{yang2022proco}
Zhixiong Yang, Junwen Pan, Yanzhan Yang, Xiaozhou Shi, Hong-Yu Zhou, Zhicheng Zhang, and Cheng Bian,
\newblock ``Proco: Prototype-aware contrastive learning for long-tailed medical image classification,''
\newblock in {\em International conference on medical image computing and computer-assisted intervention}. Springer, 2022, pp. 173--182.

\bibitem{ju2021relational}
Lie Ju, Xin Wang, Lin Wang, Tongliang Liu, Xin Zhao, Tom Drummond, Dwarikanath Mahapatra, and Zongyuan Ge,
\newblock ``Relational subsets knowledge distillation for long-tailed retinal diseases recognition,''
\newblock in {\em International Conference on Medical Image Computing and Computer-Assisted Intervention}. Springer, 2021, pp. 3--12.

\bibitem{holste2023cxr}
Gregory Holste, Song Wang, Ajay Jaiswal, Yuzhe Yang, Mingquan Lin, Yifan Peng, and Atlas Wang,
\newblock ``Cxr-lt: Multi-label long-tailed classification on chest x-rays,''
\newblock {\em PhysioNet}, vol. 5, no. 19, pp. 1, 2023.

\bibitem{lin2025cxr}
Mingquan Lin, Gregory Holste, Song Wang, Yiliang Zhou, Yishu Wei, Imon Banerjee, Pengyi Chen, Tianjie Dai, Yuexi Du, Nicha~C Dvornek, et~al.,
\newblock ``Cxr-lt 2024: A miccai challenge on long-tailed, multi-label, and zero-shot disease classification from chest x-ray,''
\newblock {\em arXiv preprint arXiv:2506.07984}, 2025.

\bibitem{bustos2020padchest}
Aurelia Bustos, Antonio Pertusa, Jose-Maria Salinas, and Maria De~La Iglesia-Vaya,
\newblock ``Padchest: A large chest x-ray image dataset with multi-label annotated reports,''
\newblock {\em Medical image analysis}, vol. 66, pp. 101797, 2020.

\bibitem{PhysioNet-mimic-cxr-2.0.0}
Alistair Johnson, Tom Pollard, Roger Mark, Seth Berkowitz, and Steven Horng,
\newblock ``{MIMIC-CXR Database},''
\newblock {\em {PhysioNet}}, Sept. 2019,
\newblock Version 2.0.0.

\bibitem{summers2019nih}
R~Summers,
\newblock ``Nih chest x-ray dataset of 14 common thorax disease categories,''
\newblock {\em NIH Clinical Center: Bethesda, MD, USA}, 2019.

\bibitem{de2025padchest}
Daniel~Coelho de~Castro, Aurelia Bustos, Shruthi Bannur, Stephanie~L Hyland, Kenza Bouzid, Maria~Teodora Wetscherek, Maria~Dolores S{\'a}nchez-Valverde, Lara Jaques-P{\'e}rez, Lourdes P{\'e}rez-Rodr{\'\i}guez, Kenji Takeda, et~al.,
\newblock ``Padchest-gr: A bilingual chest x-ray dataset for grounded radiology report generation,''
\newblock {\em NEJM AI}, vol. 2, no. 7, pp. AIdbp2401120, 2025.

\bibitem{everingham2010pascal}
Mark Everingham, Luc Van~Gool, Christopher~KI Williams, John Winn, and Andrew Zisserman,
\newblock ``The pascal visual object classes (voc) challenge,''
\newblock {\em International journal of computer vision}, vol. 88, no. 2, pp. 303--338, 2010.

\bibitem{codabench}
Zhen Xu, Sergio Escalera, Adrien Pavão, Magali Richard, Wei-Wei Tu, Quanming Yao, Huan Zhao, and Isabelle Guyon,
\newblock ``Codabench: Flexible, easy-to-use, and reproducible meta-benchmark platform,''
\newblock {\em Patterns}, vol. 3, no. 7, pp. 100543, 2022.

\end{thebibliography}

\end{document}